# Nonmonotonic Reasoning via Possibility Theory

Ronald R. Yager
Machine Intelligence Institute
Iona College
New Rochelle, N.Y. 10801

## Abstract

We introduce the operation of possibility qualification and show how this modal-like operator can be used to represent "typical" or default knowledge in a theory of nonmonotonic reasoning. We investigate the representational power of this approach by looking at a number of prototypical problems from the nonmonotonic reasoning literature. In particular we look at the so called Yale shooting problem and its relation to priority in default reasoning.

## Introduction

The construction of useful knowledge based systems requires the representation and manipulation of so called commonsense knowledge [1]. Commonsense knowledge is very often characterized by pieces of knowledge that are usually true but not necessarily always true. Many rules of thumb can be considered in this category. Stereotypical characterizations are another example of this category. Two distinct avenues exist for handling these types of knowledge. The first approach is to put this type of commonsense knowledge in to a probabilistic framework. [2,3,4]. The second approach is to consider objects typical, having the characteristic in question unless otherwise informed. This second approach has been the main avenue thus far followed by the AI community [5]. The essential feature of this second approach is <u>the assumption of a piece of knowledge without conclusive evidence of its truth</u>. Within this approach one assumes some piece of commonsense knowledge as valid if it is consistent or possible within the framework of what we already know. As Etherington [6] elegantly puts it "In commonsense reasoning assumptions are often made based upon both supporting evidence and the <u>absence of contradictory evidence.</u>"

In [7] Yager introduced a reasoning system called possibilistic reasoning. This system is rooted in the theory of approximate reasoning [8]. This system provides a set based framework for representing knowledge. We feel this system provides both the conceptual simplicity and a well developed reasoning system described as desirable by Imielinski[9]. In addition it allows for the representation of partial matching and other methods for handling imprecise information. It also allows for the inclusion of probabilistic information.[10].

In [7] Yager has suggested that we can use possibility qualification as a basis for the implementation of many different kinds of commonsense knowledge. A possibility qualified statement is

$$V \text{ is } A \text{ is possible.}$$

This statement characterizes a piece of information that says our knowledge of the value of V is such that it is <u>possible</u> (or consistent with it) to assume that V lies in the set A. Note that it doesn't specifically say V lies in A. Formally this statement gets translated into

$$V \text{ is } A^+$$

where $A^+$ is a subset of the power set of the base set X. In particular for any subset G of X

$$A^+(G) = \text{Poss}[A/G] = \text{Max}_x [A(x) \wedge G(x)]$$

Essentially $A^+$ is made up of the subsets of X which intersect, are consistent, with A.

## Representation of Commonsense Knowledge

In this section we shall investigate the representation of some primary types of commonsense knowledge by the possibilistic reasoning approach.

We shall initially consider the statement
$$\underline{\text{typically}} \text{ V is A.}$$
Within the framework of the default system of Reiter [11] the above statement gets represented as a default rule of the form
$$:M(\text{V is A}) / \text{V is A.}$$
The operator :M(F) plays a central role in Reiter's theory. It essentially means "if we have not established ¬F as true." Reiter sometimes refer's to it as a consistency operator. Thus the interpretation of "typically V is A" afforded by Reiter's default reasoning system is to say "if we have not established V is ¬A then assume V is A." It is particularly worth nothing that :M(F) is satisfied if either F is true or F is unknown. Thus we see that this operator is essentially the possibility operator which we described earlier. Armed with the set based representations of approximate reasoning we easily formalize this type of knowledge. Thus essentially we can translate
$$:M(\text{V is A}) / \text{V is A}$$
into the possibility reasoning representation
$$\text{if } \underline{\text{V is A is possible}} \text{ then V is A.}$$
Using our translation rules we get
$$\text{if V is } A^+ \text{ then V is A}$$
This translates into
$$\text{V is } \neg(A^+) \cup A.$$
We shall denote $\neg(A^+)$ as $A^*$, hence we get
$$\text{V is } (A^* \cup A).$$
Furthermore assume that our knowledge base consists simply of the fact that
$$\text{V is B.}$$
Combining this with our typical knowledge we get
$$\text{V is D}$$
where $D = (A^* \cap B) \cup (A \cap B)$.
Since $A^*$ is a subset of the power set of X and B is a subset of X we convert B into a subset of the power set, thus
$$B = \{1/B\}.$$
Furthermore as discussed in [7,10]
$$A^* \cap B = \{(1 - \text{Poss}[A/B]) / B\}$$
which in turn can be expressed as a subset of X, M, where
$$M(x) = B(x) \wedge (1-\text{Poss}[A/B])$$
thus
$$D(x) = (B(x) \wedge (1-\text{Poss}[A/B]) \vee (A(x) \wedge B(x)).$$

Two extremal cases should be noted. If our typical value A is completely inconsistent with our known value, $A \cap B = \Phi$, then $\text{Poss}[A/B] = 0$ and thus
$$D(x) = B(x)$$
and hence
$$\text{V is B}$$
Thus in this case we have discounted our typical information when it conflicted with our knowledge-base.

369

On the other hand if A has some consistency with B, $A \cap B \neq \Phi$, thus Poss[A/B] = 1 then we get
$$D(x) = B(x) \wedge A(x)$$
and hence
$$V \text{ is } A \cap B.$$
Thus when our typical knowledge doesn't contradict our firm knowledge we conjunct these sources of knowledge.
In the special case when B is unknown, B = X, then we get
$$V \text{ is } A.$$

## Complementary Default Rules

An important class of default type rules are those in which we have two complementary propositions. An example of this situation is
> typically Quakers are pacifists
> typically Republicans are non-pacifists

More formally these rules are characterized by
> $P_1$: typically if V is A then U is B
> $P_2$: typically if W is C then U is $B^-$.

A coherent reasoning system should report U is B if all we know is V is A. If all we know is V is C then we should infer U is $B^-$. If we know neither A nor C or both A and C then we should report unknown for U. Let us see how our structure handles this kind of knowledge.

We represent $P_1$ as If V is A and U is B is possible then U is B similarly for $P_2$ If W is C and U is $B^-$ is possible then U is $B^-$. Formally we get
> $P_1$: $A^- \cup B^* \cup B$     (where $B^*$ = not(B is possible)
> $P_2$: $C^- \cup B^{\bullet} \cup B^-$     (where $B^{\bullet}$ = not(not B is possible)

Conjuncting these pieces of information gives us
$$H = A^- \cap C^- \cup C^- \cap B^* \cup C^- \cap B \cup B^{\bullet} \cap A^- \cup B^{\bullet} \cap B^*$$
$$\cup B^{\bullet} \cap B \cup B^- \cap A^- \cup B^- \cap B^*.$$

We let X, Y and Z denote the base sets of A, B and C respectively. Furthermore we note that $B^{\bullet} \cap B^* = \Phi$.

In the case where the value of V and W are unknown our knowledge simply consists of H. Hence U is G where $G(y) = \text{Max}_{x,z}(H(x,z,z))$. However since $\text{Max}_{z,x}[C^-(z) \wedge A^-(x)] = 1$ it follows G(y) = 1 and thus U is unknown.

If V is A and W is unknown we conjunct H with A and get
$$H' = (C^- \cap B \cap A) \cup (B^{\bullet} \cap B \cap A) \cup (B^* \cap A \cap (B^- \cup C^-)$$
From this we can infer that U is G' where
$$G'(y) = B(y) \vee (B(y) \wedge (1-\text{Poss}[B^-/B])) \vee (Y(y) \wedge (1-\text{Poss}(B/Y)))$$
which results in $G'(y) = B(y) \vee B(y) \vee 0 = B(y)$. Thus in the case where we know V is A we correctly infer U is B.

It can be easily shown, in a symmetric manner, if all we know is that W is C that we infer U is $B^-$.

The final case corresponds to the situation in which we know both V is A and W is C. In this case we conjunct $A \cap C$ with H to get H"
$$= B^{\bullet} \cap B \cup B^- \cap B^*.$$
Projecting onto Y we get $\text{Proj}_Y H" = B(y) \cup B^-(y) = 1$, hence in this case we get U is "unknown" Thus we see that our system makes the correct inference in the face of complimentary default rules.

## Priority in Default Rules

In [12] Hanks and McDermott introduced a problem which raised a number of significant issues for the use of non-monotonic reasoning schemes. In [13] Pearl discusses a modified version of this problem which he aptly calls "the Yale shooting problem." We shall investigate this modified version of the Hanks-McDermott problem suggested by Pearl as an example of use of our methodology as well as a vehicle to introduce a number of meta-rules useful in providing priority information for non-monotonic type knowledge.

Assume we have a knowledge base which consists of the following four pieces of commonsense knowledge:
- $D_1$: typically if a gun is loaded at time 1 it is loaded at time $t_2$
- $D_2$: typically if a person is alive at time $t_2$ he is alive at $t_3$
- $D_3$: typically if a person is alive at time $t_2$ and shot at $t_2$ and the gun is loaded at $t_2$ then their not alive at $t_3$
- $D_4$: typically if a person is alive at time $t_2$ and shot at $t_2$ with a gun that is unloaded at $t_2$ then they are alive at $t_3$

We shall use the following notation
$A_1$ - alive at time $t_1$; $L_1$ - gun loaded at $t_1$; $S_1$ - gun shot at $t_1$
Using our notation we have the following representation of out knowledge

$D_1$: $L_1$ and $L_2$ <u>is possible</u> then $L_2$
$D_2$: $A_2$ and $A_3$ <u>is possible</u> then $A_3$
$D_3$: $A_2$ and $S_2$ and $L_2$ and $A_3$ <u>is possible</u> then $A^-_3$
$D_4$: $A_2$ and $S_2$ and not $L_2$ and $A_3$ <u>is possible</u> then $A_3$

This can be formally expressed as

$D_1$: $L^-_1 \cup L_2^* \cup L_2$
$D_2$: $A^-_2 \cup A_3^* \cup A_3$
$D_3$: $A^-_2 \cup S^-_2 \; L^-_2 \cup A_3^* \cup A^-_3$
$D_4$: $A^-_2 \cup S^-_2 \cup L_2 \cup A_3^* \cup A_3$

In addition to these four default rules we have the following three pieces of factual knowledge $L_1$, $A_2$ and $S_2$.

We now introduce two other pieces of knowledge. These are two pieces of <u>meta-knowledge</u> with respect to priorities on the introduction of default rules

    MR-1: (1) specialization priority
    MR-2: (2) temporal priority

We shall first discuss in turn these two meta priority rules.
Assume we have two default rules:
    $R_1$: typically (if A then E)
    $R_2$: typically (if A and B then G)

The meta-rule of <u>specialization priority</u> says that $R_2$ has priority over $R_1$, it is introduced first. In a simplistic way we see for this is that $R_2$ has less potential exceptions than $R_1$, since B is not an exception.

The second rule, temporal priority, is closely related to Shoham's concept of chronological ignorance [14] as well as the basic idea of causality. It is a principle useful for knowledge (default rules) that have a predictive nature to them. The essential idea responsible for this meta priority rule is an assumed causality in the world. The basic idea here is that anything that happens in the world at time $t_1$ must be caused by things that happened in the world at times before $t_j$, $t \leq t_j$. Thus this requires objects in the antecedent of predictive rule happen before the consequent. Essentially this principle states that if

    $R_1$: Typically $[R_1 \; f(t_1) \to f(t_2)]$
    $R_2$: Typically $[R_2 \; f(t_3) \to f(t_4)]$

and if $t_2 < t_4$ then $R_1$ can be introduced prior to $R_2$.

371

Having discussed these meta rules we are now in a position to make the appropriate inferences based upon our knowledge base: commonsense knowledge, facts and meta-priority rules.

The following analysis sets up the appropriate priority schedule. We start with seven pieces of data
$$F_1, F_2, F_3, D_1, D_2, D_3, D_4$$
Because $F_1, F_2, F_3$ are facts not default knowledge the D's get introduced after the F's hence we get
$$\{F_1, F_2, F_3\}$$
$$\text{prior to}$$
$$\{D_1, D_2, D_3, D_4\}$$
Because of the principle of temporal priority, $D_1$ has an earlier time in its consequent, $D_2, D_3$ and $D_4$ get introduced after $D_1$. Thus we have
$$\{F_1, F_2, F_3\}$$
$$\text{prior to}$$
$$\{D_1\}$$
$$\text{prior to}$$
$$\{D_2, D_3, D_4\}$$
Because of the principle specialization priority $D_2$ gets introduced after $D_3$ and $D_4$. Thus our final priority schedule is
$$\{F_1, F_2, F_3\}$$
$$\text{prior to}$$
$$\{D_1\}$$
$$\text{prior to}$$
$$\{D_3, D_4\}$$
$$\text{prior to}$$
$$\{D_2\}$$

We start by simultaneously introducing the facts $L_1$, $A_2$ and $S_2$ thus letting H be a knowledge we get
$$H = L_1 \cap A_2 \cap S_2$$
Next we introduce $D_1$, thus we get
$$H = L_1 \cap A_2 \cap S_2 \cap (L^-_1 \cup L^*_2 \cup L_2)$$
$$H = A_2 \cap S_2 \cap L_1 \cap (L^*_2 \cup L_2)$$
We effect the rule $D_1$ by effecting the second order knowledge $L^*_2$, since $L_2$ is possible we obtain
$$H = A_2 \cap S_2 \cap L_1 \cap L_2$$
We next simultaneously introduce $D_3$ and $D_4$
$$H = (A_2 \cap S_2 \cap L_2 \cap L_1) \cap (A^-_2 \cup S^-_2 \cup L_2 \cup A^*_3 \cup A_3)$$
$$\cap (A^-_2 \cup S^-_2 \cup L^-_2 \cup A^*_3 \cup A_3).$$
Hence
$$H = A_2 \cap S_2 \cap L_2 \cap L_1 \cap [L_2 \cup A^*_3 \cup A_3] \cap [A^*_3 \cup A^-_3]$$
$$H = A_2 \cap L_2 \cap L_1 \cap S_2 \cap [L_2 \cap A^*_3 \cup L_2 \cap A^-_3 \cup A^*_3 \cap A^*_3$$
$$\cup A^*_3 \cap A^-_3 \cup A_3 \cap A^*_3]$$
$$H = A_2 \cap L_2 \cap L_1 \cap S_2 \cap (A^*_3 \cup A^-_3)$$
Effecting $A^*_3$, since there is nothing stopping $A^-_3$ we get
$$H = S_2 \cap L_2 \cap L_1 \cap S_2 \cap [A^*_3 \cup A^-_3]$$
Finally introducing $D_2$, we get
$$H = S_2 \cap L_2 \cap A_2 \cap A^-_3 \cap L_1 \cap (A^-_2 \cup A^*_3 \cup A_3)$$
$$H = S_2 \cap L_2 \cap A_2 \cap A^-_3 \cap L_1 \cap A^*_3$$
Since $A^*_3 = \{A^-_3, \emptyset\}$ its intersection with $A^-_3$ is $A^-_3$ we get that the final version of H after the inclusion of all our knowledge is
$$H = S_2 \cap L_2 \cap A_2 \cap A^-_3 \cap L_1.$$
From this it follows that $A^-_3$ is true. Hence our victim is not alive at time 3.

We see that the use of these meta-priority rules along with our possibilistic reasoning process leads to the correct results in this litmus test situation.

## References

(1) McCarthy, J., "Epistemological problems in artificial intelligence," Proc. Int. Joint Conf. on A.I., Cambridge, Mass., 223-227, 1977

(2) Rich, E., "Default reasoning as likelihood reasoning," Proc. Amer. Assoc. for A.I., Washington, D.C., 348-351, 1983

(3) Zadeh, L.A., "Syllogistic reasoning in fuzzy logic and its application to usuality and reasoning with dispositions," IEEE Trans. on Systems, Man and Cybernetics 15, 754-763, 1985

(4) Yager, R.R., "On implementing usual values," Proc. Second Workshop on Uncertainty in Artificial Intelligence, Philadelphia, 339-346, 1986

(5) Reiter, R., "Nonmonotonic reasoning," Annual Reviews of Computer Science 2, 81-132, 1987.

(6) Etherington, D.W., "Formalizing nonmonotonic reasoning systems," Artificial Intelligence 31, 41-85, 1987

(7) Yager, R.R., "Using approximate reasoning to represent default knowledge," Artificial Intelligence 31, 99-112, 1987

(8) Zadeh, L.A., "A theory of approximate reasoning," in Machine Intelligence vol. 9, Hayes, J.E., Michie, D. & Kulich, L.I. (Eds.), 149-194, John Wiley & Sons: New York, 1979

(9) Imielinski, T., "Results on translating defaults to circumscription," Artificial Intelligence 32, 131-146, 1987

(10) Yager, R.R., "Probabilistic qualification and default rules," in Uncertainty in Knowledge Based Systems, Bouchon, B. & Yager, R.R., Springer-Verlag: Berlin, 41-57, 1987.

(11) Reiter, R., "On reasoning by default," Proc. Second Symp. on Theoretical Issues in Natural Language Processing, Urbana, 1978

(12) Hanks, S. & McDermott, D., "Default reasoning, nonmonotonic logics and the frame problem," Proc AAAI-86, Philadelphia, 328-333, 1986

(13) Pearl, J. "A probabilistic treatment of the Yale shooting problem," Tech. Report CSD-8700XX, R-100, UCLA, 1987

(14) Shoham, Y., "Chronological ignorance: time, nonmonotonicity, necessity and causal theories," Proc. AAAI-86, Philadelphia, 389-393, 1986